\documentclass[11pt]{article}

\usepackage[final]{acl}

\usepackage{times}
\usepackage{latexsym}

\usepackage[T1]{fontenc}
\usepackage[utf8]{inputenc}

\usepackage{microtype}

\usepackage{inconsolata}
\usepackage{algorithm}
\usepackage{amsmath}
\usepackage{xcolor}
\usepackage{booktabs}  
\usepackage{multirow} 
\usepackage{graphicx} 
\usepackage[table]{xcolor} 
\usepackage{arydshln}    
\usepackage{algorithm}
\usepackage{algpseudocode}

\usepackage{newfloat}
\usepackage{listings}
\usepackage{graphicx}

\title{MIND Your Reasoning: A Meta-Cognitive Intuitive-Reflective Network for Dual-Reasoning in Multimodal Stance Detection}

\author{
 \textbf{Bingbing Wang\textsuperscript{1,2}},
 \textbf{Zhengda Jin\textsuperscript{1}},
 \textbf{Bin Liang\textsuperscript{3}},
 \textbf{Wenjie Li\textsuperscript{2}},
\\
 \textbf{Jing Li\textsuperscript{2*}},
 \textbf{Ruifeng Xu\textsuperscript{1*}},
  \textbf{Min Zhang\textsuperscript{1}},
\\
 \textsuperscript{1}Harbin Institute of Technology, Shenzhen, China, \\
 \textsuperscript{2}The Hong Kong Polytechnic University, Hong Kong, China,\\
 \textsuperscript{3}The Chinese University of Hong Kong, Hong Kong, China
\\
}

\begin{document}
\maketitle
\begin{abstract}
Multimodal Stance Detection (MSD) is a crucial task for understanding public opinion on social media. 
Existing methods predominantly operate by learning to fuse modalities. They lack an explicit reasoning process to discern how inter-modal dynamics, such as irony or conflict, collectively shape the user's final stance, leading to frequent misjudgments. 
To address this, we advocate for a paradigm shift from \textit{learning to fuse} to \textit{learning to reason}. 
We introduce \textbf{MIND}, a \textbf{M}eta-cognitive \textbf{I}ntuitive-reflective \textbf{N}etwork for \textbf{D}ual-reasoning.
Inspired by the dual-process theory of human cognition, 
MIND operationalizes a self-improving loop. It first generates a rapid, intuitive hypothesis by querying evolving Modality and Semantic Experience Pools. Subsequently, a meta-cognitive reflective stage uses Modality-CoT and Semantic-CoT to scrutinize this initial judgment, distill superior adaptive strategies, and evolve the experience pools themselves.
These dual experience structures are continuously refined during training and recalled at inference to guide robust and context-aware stance decisions. 
Extensive experiments on the MMSD benchmark demonstrate that our MIND significantly outperforms most baseline models and exhibits strong generalization.
\end{abstract}

\section{Introduction}
The proliferation of social platforms supporting multimodal content, such as text or images \cite{8-dar2025social}, has made Multimodal Stance Detection (MSD) a critical task for understanding public opinion \cite{1-liang2024multi,2-niu2024multimodal}. MSD aims to identify the expressed attitude 
toward a specific target by integrating both textual and visual information, enabling a more comprehensive analysis \cite{6-liang2022zero}. This task applies widely to political discourse analysis, misinformation detection, and public health communication, making it a crucial tool for decoding complex online interactions \cite{2-niu2024multimodal,3-pangtey2025large}.

\begin{figure}[!t]
  \centering
  \includegraphics[width=\linewidth]{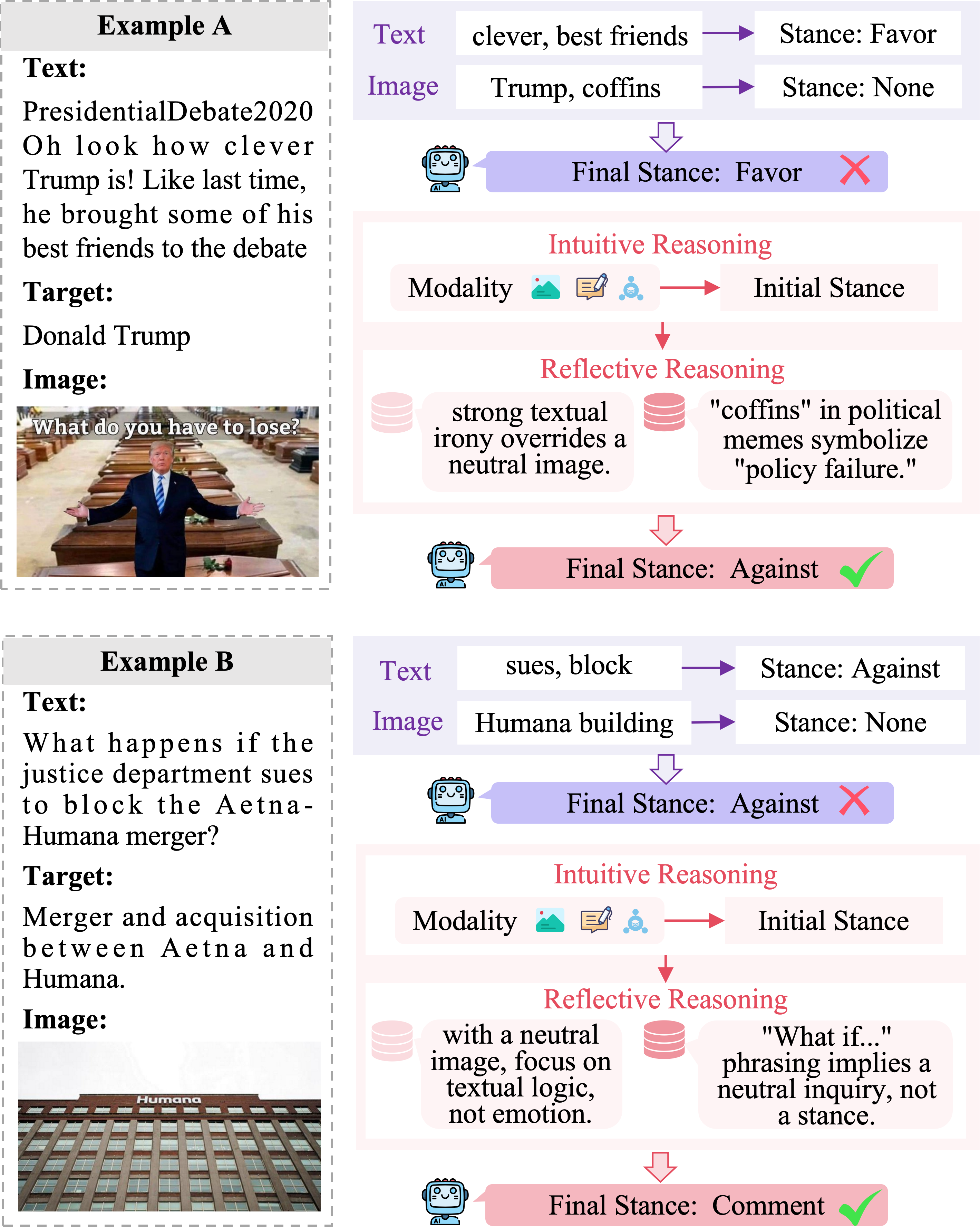}
  \caption{Two examples contrasting existing methods (learning to fuse) in purple with our method (learning to reason) in pink.}
  \label{f1}
  \vspace{-0.3cm}
\end{figure}

While recent methods have advanced from static fusion to dynamic mechanisms like cross-attention, they remain fundamentally constrained by a paradigm of learning to fuse \cite{1-liang2024multi, 31-zhang2025logic}.
This paradigm excels at weighing feature importance but lacks the explicit, high-level reasoning process needed to interpret inter-modal stance expressions, especially in \textbf{semantic traps} where seemingly consistent modalities convey a contradictory stance through nuanced cues like irony or cultural context.
These traps manifest in at least two critical ways. First, models are easily deceived by cross-modal irony, as seen in Figure \ref{f1}. A fusion-based model would likely interpret the positive text and the image objects as evidence for a FAVOR stance. It fails because it cannot perform the crucial reasoning step: recognizing that the text's "best friends" sarcastically refers to the image's "coffins," a grim symbol requiring world knowledge to interpret as a biting critique. 
Second, the challenge extends to intra-modal logical fallacies. A text can deceive a model focused on keywords like "sues" and "block" to conclude an AGAINST stance, completely missing that the "What if..." framing makes it a neutral COMMENT.
In both scenarios, the failure stems not from an inability to weigh features, but from the absence of a high-level reasoning process to interpret non-literal intent and logical structure.

To bridge this gap, we advocate for a paradigm shift from learning to fuse to learning to reason. We introduce and pioneer \textbf{MIND}, a \textbf{M}eta-cognitive \textbf{I}ntuitive-reflective \textbf{N}etwork for \textbf{D}ual-reasoning that realizes a more human-like, adaptive reasoning process for a complex multimodal stance detection. 
Inspired by the dual-process theory, MIND synergizes experience-driven intuitive reasoning with meta-cognitive reflective reasoning.
First, an intuitive stage queries the Modality (MEP) and Semantic (SEP) Experience Pools to form an initial hypothesis from past reasoning patterns. This hypothesis is then scrutinized by a meta-cognitive reflective stage, where a Modality Chain of Thought (Modality-CoT) resolves inter-modal dynamics and formulates adaptive fusion strategies, while a complementary Semantic-CoT ensures a robust final decision. 
Extensive experiments on the public MMSD benchmark demonstrate that MIND significantly outperforms baseline models and exhibits strong generalization, validating the effectiveness of our meta-cognitive approach.
Our main contributions are as follows:

\begin{itemize}
    \item We pioneer a paradigm shift for MSD from learning to fuse to learning to reason. Our dual-reasoning framework is the first to explicitly model a meta-cognitive process for resolving complex stance expressions.
    \item We operationalize this paradigm with a novel self-improving meta-cognitive loop, where dual Experience Pools store and evolve explicit reasoning strategies that are then refined by a reflective Chain-of-Thought process.
    \item Experiments conducted on the public dataset demonstrate the superiority of our MIND, which significantly outperforms most baselines and exhibits strong generalization. 
\end{itemize}

\begin{figure*}[!t]
  \centering
  \includegraphics[width=\linewidth]{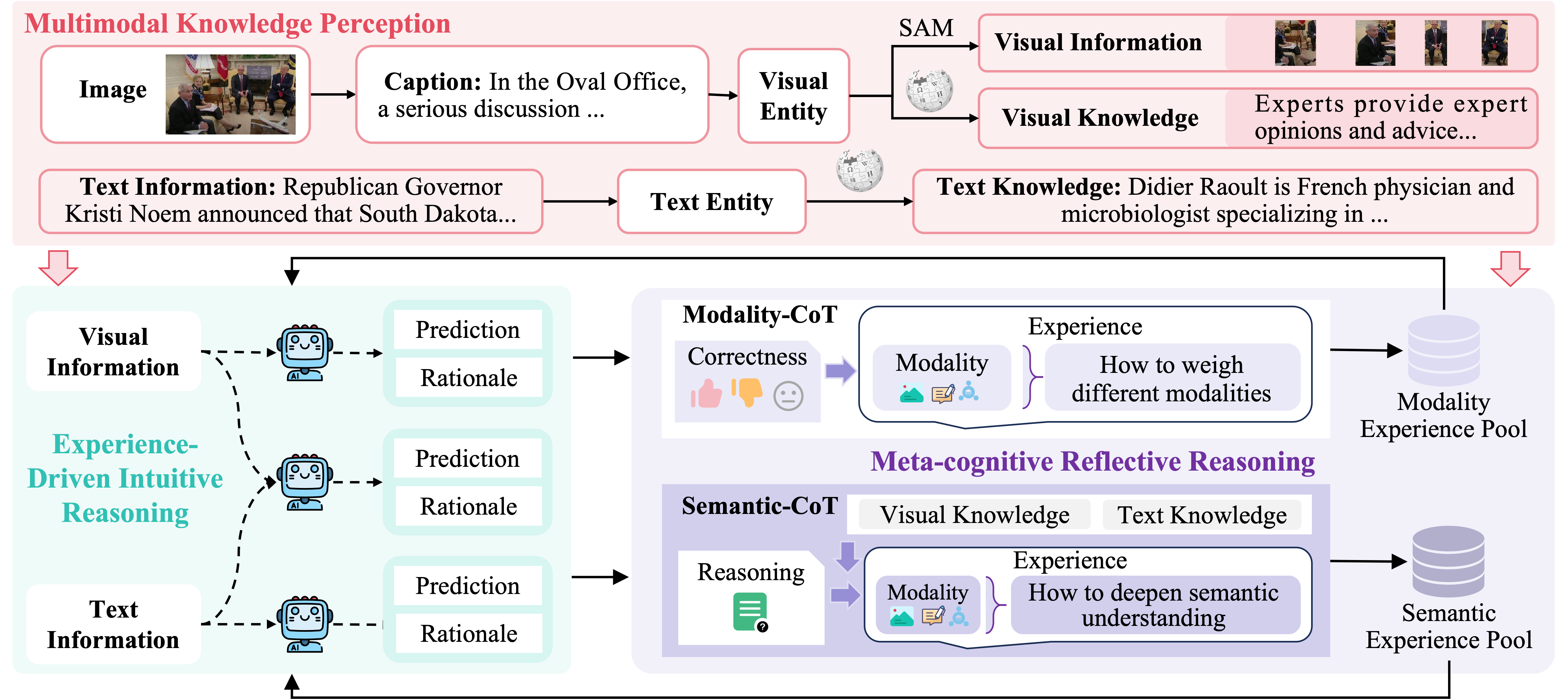}
  \caption{Overflow of our MIND framework for multimodal stance detection.}
  \label{f2}
\end{figure*}

\section{Related Works}
\subsection{LLM-based Stance Detection}
The design of modern stance detection frameworks has been significantly influenced by LLMs. Early approaches focused on harnessing their zero-shot and few-shot capabilities, using direct prompting and Chain-of-Thought (CoT) to elicit stance predictions without labeled data or fine-tuning \cite{13-zhang2023investigating, 14-zhang2022would}. Other works have pursued fine-tuning by fusing retrieved background knowledge with input pairs to improve learning through backpropagation \cite{11-li2023stance}.
Subsequent research has sought to further unleash LLM reasoning. Innovations include integrating multi-perspective insights via CoT and prompt-tuning \cite{15-ding2024cross}, formalizing reasoning with first-order logic rules for enhanced interpretability \cite{16-zhang2024knowledge}, and architecting multi-agent systems where LLMs collaborate for detection \cite{12-lan2024stance, 19-wang2024deem}. 
However, these methods are primarily developed for pure textual stance detection and have not yet explored the integration of multimodal information such as images.

\subsection{Multimodal Stance Detection}

The proliferation of social media has driven the rise of multimodal content (text, image) in forms like memes and reels \cite{8-dar2025social,9-bansal2024hybrid}, which convey implicit opinions through diverse signals, posing challenges for stance detection. Modeling semantically heterogeneous cross-modal interactions remains difficult for traditional models, whereas transformer architectures and LLMs excel here via self-attention and transfer learning.
Recent approaches improved stance classification through multimodal fusion. For example, 
\citet{2-niu2024multimodal} proposed MmMtCSD and MLLM-SD for text-image pairs in multi-turn dialogues. \citet{23-kuo2024advancing} fused textual, visual, and social signals in political discourse using S-BERT, BEiT, and GraphSAGE. \citet{24-xie2021sern} enhanced fake news detection via sentence-guided visual attention. \citet{25-khiabani2023few} and \citet{26-barel2025acquired} addressed few-shot or noisy settings with graph-based cues. \citet{27-yao2023end} extended MSD to fact-checking via web evidence integration. However, these approaches typically employ rigid fusion strategies, overlooking the dynamic interplay and context-dependent reliability of modalities, which makes them vulnerable to noisy or contradictory information.

\section{Method}
This section introduces our novel MIND framework for MSD in detail. Drawing inspiration from the dual-process theory of human cognition \cite{47-kahneman2011fast}, a psychological model that posits two distinct modes of thought. One is an intuitive, fast `System 1' that relies on heuristics and prior experiences to make rapid judgments. The other is a deliberate, slow `System 2' that engages in analytical reasoning to refine and validate these initial assessments. 
Our MIND framework operationalizes this cognitive paradigm for MSD.

Given a text-image pair $(u, v)$, the aim of MIND is to identify the stance label $y$,  towards a specific target $t$.
As depicted in Figure~\ref{f2}, our MIND consists of three primary components: 
1) \textbf{Multimodal Knowledge Perception}, which extracts comprehensive multimodal information and captures relevant external knowledge. 
2) \textbf{Experience-Driven Intuitive Reasoning} (System 1), which queries the experience pools to form an initial stance hypothesis; and 
3) \textbf{Meta-cognitive Reflective Reasoning} (System 2), which employs complementary reasoning chains to scrutinize the initial hypothesis and continuously refine the dual experience pools.

\subsection{Multimodal Knowledge Perception}
\label{sec:knowledge_perception}

This initial stage constructs a rich semantic foundation by extracting and refining knowledge from both textual and visual modalities. The goal is to capture not only high-level conceptual information but also grounded, fine-grained evidence, which is crucial for nuanced reasoning.

\subsubsection{Textual Knowledge Attainment}
For the text input $u$, we perform a three-step knowledge extraction process. First, we employ a Multimodal Large Language Model (MLLM) with strategically designed prompts to identify a set of key semantic entities, denoted as $\mathcal{E}_u$. To ground these entities in external world knowledge, we retrieve corresponding background information for each entity $e \in \mathcal{E}_u$ from the Wikipedia. As the raw retrieved knowledge is often verbose and contains noisy information, we leverage the MLLM again to distill it into a concise, semantically-rich summary. This final summary, denoted as the text knowledge $\mathcal{K}_u$, serves as the knowledge for the text modality.

\subsubsection{Visual Information Grounding}
For the image input $v$, we capture both high-level semantic concepts and low-level visual information.
First, to extract semantic concepts, an MLLM generates an image caption, identifies a set of visual entities $\mathcal{E}_v$ from the caption. These entities are then used to query Wikipedia, and the resulting information is summarized into a condensed visual knowledge $\mathcal{K}_v$.
However, knowledge alone is insufficient to capture the visual manifestation of these concepts. To address this, we ground the semantic entities in the image's pixel space. We leverage the identified entity set $\mathcal{E}_v$ to guide the open-vocabulary segmentation model SAM~\cite{29-ravi2024sam} in precisely segmenting the image regions corresponding to each entity, yielding a set of entity-aligned visual information $\mathcal{V}$ where each element is a segmented image directly showcasing the visual appearance of a specific semantic entity.
\begin{equation}
    \mathcal{V} = \{\bar  v_i\}^N_{i=1} = \text{SegModel}(v, \mathcal{E}_v) 
    \label{eq:visual_refined}
\end{equation}
where $N$ indicates the number of sub-images.

\subsection{Experience-Driven Intuitive Reasoning}
This stage simulates rapid, intuitive judgment by leveraging experience pools accumulated during training to quickly formulate an initial stance hypothesis. The process consists of two main steps: modal synergistic experience retrieval and experience-driven stance prediction.

\subsubsection{Modal Synergistic Experience Retrieval}
To retrieve relevant past experiences, we introduce a modal synergistic retrieval mechanism. Unlike conventional text-only systems, our approach leverages both text and visual information to query the modality and semantic experience pool.
Each pool stores a collection of experiences $\{E_j\}$, where each experience is a key-value pair. The key is a pre-computed bimodal embedding $(k_u^j,k_\mathcal{V}^j)$, representing the experience's core features, while the value contains the associated experiential data. 

The retrieval process commences by encoding the current text input text $u$ and sub-images $\mathcal{V}$ into their respective query vectors, $q_u$ and $q_\mathcal{V}$:
\begin{equation}
% \begin{aligned}
    q_{u} = \text{Enc}(u), ~q_{\mathcal{V}} = \frac{1}{N} \sum^N_{i=1} \text{Enc}(\bar v_i)
% \end{aligned}
\end{equation}
where $q_u$ and $q_\mathcal{V}$ are produced by the BGE-VL-CLIP $\text{Enc}$ \cite{zhou2025megapairs}.
The stored key vectors $k_u^j$ and $k^j_\mathcal{V}$ for each Experience $E_j$ are computed analogously as $q_u$ and $q_\mathcal{V}$.
With the query vectors established, we measure their similarity to the key vectors of each stored experience $E_j$ by computing unimodal relevance scores for text $S_u$ and vision $S_v$ using cosine similarity:
\begin{equation}\small
% \begin{aligned}
    S_{u}(q_u,k_u^j) = \frac{q_u \cdot k_u^j}{\|q_u\|\|k_u^j\|}, ~ S_{\mathcal{V}}(q_\mathcal{V},k_\mathcal{V}^j) = \frac{q_\mathcal{V} \cdot k_\mathcal{V}^j}{\|q_\mathcal{V}\|\|k_\mathcal{V}^j\|} 
% \end{aligned}
\end{equation}

To obtain a holistic measure of relevance, we fuse these scores into a unified Modality Relevance Score (MRS), denoted as $S(E_j)$, via a weighted linear combination:
\begin{equation}
    S(E_j) = \alpha \cdot S_u(q_u,E_j)+(1-\alpha) \cdot S_\mathcal{V}(q_\mathcal{V},E_j)
\end{equation}
where $\alpha \in [0,1]$ balances the weight of each modality for retrieval.
Ultimately, experiences are ranked in descending order based on their MRS. The top-$k$ experiences with scores excedding a relevance threshold $\tau$ are then selected, yielding the modality experience set $\{\mathcal{S}_i^{\text{ME}}\}^k_{i=1}$ and the semantic experience set $\{\mathcal{S}_i^{\text{SE}}\}^k_{i=1}$.

\subsubsection{Experience-driven Stance Prediction}
Following the retrieval of relevant past experiences, our MIND agent $A_R$ proceeds to generate initial, intuitive stance hypotheses. We formulate a comprehensive prompt $\mathcal{P}$ that integrates two key components: (1) text information $u$ and visual information $\mathcal{V}$; (2) the retrieved experience sets $\{\mathcal{S}^{\text{ME}}\}^k_{i=1}$ and $\{\mathcal{S}^{\text{SE}}\}^k_{i=1}$, which serve as in-context exemplars. This prompt instructs $A_R$ to perform a preliminary analysis from three distinct perspectives: text-only, vision-only, and combined multimodal. For each perspective, the agent generates an initial stance prediction $\hat y$ and a corresponding rationale $r$. This process is formally expressed as:
\begin{equation}
    (\hat y_c,r_c)=A_R(\mathcal{P}(c,\{\mathcal{S}^{\text{ME}}\}^k_{i=1},\{\mathcal{S}^{\text{SE}}\}^k_{i=1}))
    %~\text{for}~c \in \{u,\mathcal{V},(u,\mathcal{V})\}
\end{equation}
where $c \in \{u,\mathcal{V},(u,\mathcal{V})\}$ denotes the context of evaluation: the text information $u$, the visual information $\mathcal{V}$ or the full multimodal pair $(u,\mathcal{V})$. 
These generated pairs $(\hat y_c,r_c)$ serve as essential intermediate reasoning traces that are scrutinized and refined during the subsequent reflective reasoning stage.

\subsection{Meta-cognitive Reflective Reasoning}
Meta-cognitive Reflective Reasoning is analogous to the `System 2' slow thinking process in dual-process cognitive theory and serves as the refinement module of our framework. This process is governed by experience formulation, where novel insights are generated through structured reflection, and experience evolution, where the experience pools are refined by integrating these new insights.

\subsubsection{Experiences Formulation}
The primary objective of experience formulation is to facilitate meta-cognitive reflection, enabling the agent to distill actionable experience from its performance outcomes. To this end, the MIND agent employs a structured, multi-step reasoning process, analogous to a CoT, to generate new insights.

To generate the modality experience, the Modality-CoT initiates a self-diagnostic process. It evaluates the accuracy of the initial predictions for text $\hat y_u$, vision $\hat y_\mathcal{V}$, and combined multimodal input $\hat y_{(u,\mathcal{V})}$, along with their corresponding rationales $r_{u}$, $r_{\mathcal{V}}$, and $r_{(u,\mathcal{V})}$.
By analyzing the failures and successes of these initial judgments, it distills a high-level \textit{modality insight} $\mathcal{I}_m$. This insight is not merely a summary, but an adaptive fusion strategy that dictates how to dynamically balance unimodal and multimodal cues in future reasoning.

In parallel, the Semantic-CoT is tasked with forming the \textbf{semantic experience}. This process involves a deep interrogation of the initial reasoning's semantic grounding. It synthesizes the agent's internal rationales with the externally-grounded textual knowledge $\mathcal{K}_u$ and visual knowledge $\mathcal{K}_v$. 
This synthesis allows the agent to identify and rectify content-level biases or shallow interpretations. The outcome is a semantic insight $\mathcal{I}_s$, an actionable directive that refines the model's strategies for achieving deeper semantic and contextual understanding.

\subsubsection{Experiences Evolution}
Upon deriving new insights $\mathcal{I}_m$ and $\mathcal{I}_s$, MIND updates its experience pools to consolidate and refine existing experiences by first employing the modal synergistic retrieval mechanism to retrieve the top-$k$ most relevant prior experiences from each pool. These experiences, denoted as $\{ \mathcal{\bar S}_i^{\text{ME}}\}^k_{i=1}$ from MEP and $\{ \mathcal{\bar S}_i^{\text{SE}}\}^k_{i=1}$ from SEP, have scores exceeding a relevance threshold $\tau$ and are selected based on the current text information $u$ and visual information $\mathcal{V}$. 
If no experiences meet this threshold, the new insights are encapsulated as new key-value entries and appended to their respective pools.  Conversely, MIND performs targeted updates on these entries to prevent redundant memories.

This evolutionary update hinges on a fusion operation. For each retrieved modality experience, MIND initiates a re-reasoning process, using a prompt to synthesize the prior experience $\{\mathcal{\bar S}^{\text{ME}}_{i}\}^k_{i=1}$ with the new modality insight $I_m$. This generates a more comprehensive, updated modality experience pool $\mathcal{\hat S}^{\text{ME}}$,  which then overwrites the original entry.
A parallel process is executed for the semantic experiences, updating $\{ \mathcal{\bar S}_i^{\text{SE}}\}^k_{i=1}$ with $I_s$ to produce new semantic experience pool $\mathcal{\hat S}_i^{\text{SE}}$. This dual-update ensures both experience pools remain synchronized and continuously refined. In essence, this mechanism enables MIND to build a compact and coherent knowledge base through continuous refinement, supporting efficient learning.

\begin{table*}[t!]
\centering
\small
\setlength{\tabcolsep}{2.5pt}
\resizebox{\textwidth}{!}{%
\begin{tabular}{ll ccccccccccccc}
\toprule
\multirow{2.5}{*}{MODALITY} & \multirow{2.5}{*}{METHOD} & \multicolumn{2}{c}{MTSE} & \multicolumn{1}{c}{MCCQ} & \multicolumn{5}{c}{MWTWT} & \multicolumn{2}{c}{MRUC} & \multicolumn{2}{c}{MTWQ} \\
\cmidrule(lr){3-4} \cmidrule(lr){5-5} \cmidrule(lr){6-10} \cmidrule(lr){11-12} \cmidrule(lr){13-14}
& & DT & JB & CQ & CA & CE & AC & AH & DF & RUS & UKR & MOC & TOC \\
\midrule
\multirow{5}{*}{Text-only} & 
BERT %\cite{33-devlin2019bert}
& 48.25 & 52.04 & 66.57 & 75.62 & 60.85 & 63.05 & 59.24 & 81.53 & 41.25 & 46.80 & 57.77 & 45.91 \\
& RoBERTa %\cite{34-liu2019roberta}
& 58.39 & 60.79 & 66.57 & 69.56 & 65.03 & 69.74 & 67.99 & 79.21 & 39.52 & 57.66 & 55.22 & 48.88 \\
& KEBERT %\cite{35-kawintiranon2022polibertweet}
& 64.50 & 69.81 & 66.84 & 71.67 & 67.56 & 69.29 & 69.74 & 80.57 & 41.55 & 59.01 & 58.15 & 47.75 \\
\cdashline{2-14}
& LLaMA2 %\cite{36-touvron2023llama}
& 53.23 & 52.67 & 47.40 & 34.89 & 41.95 & 49.09 & 44.32 & 30.21 & 38.84 & 38.54 & 55.31 & 46.51 \\
& GPT4%\footnote{https://openai.com/research/gpt-4} 
& 68.74 & 66.39 & 65.84 & 63.14 & 65.12 & \underline{69.93} & \textbf{71.62} & 52.69 & 41.64 & 53.76 & 58.05 & 49.81 \\
\midrule
\multirow{3}{*}{Vision-only} & ResNet %\cite{37-he2016deep}
& 37.89 & 38.59 & 47.16 & 39.89 & 42.20 & 43.52 & 37.05 & 50.34 & 35.10 & 40.00 & 42.02 & 33.94 \\
& ViT %\cite{38-dosovitskiy2020image}
& 40.48 & 40.42 & 46.64 & 46.63 & 50.00 & 40.16 & 46.32 & 50.86 & 33.31 & 39.87 & 38.63 & 35.53 \\
& SwinT %\cite{39-liu2021swin}
& 39.89 & 40.43 & 48.80 & 46.30 & 46.99 & 41.02 & 47.39 & 51.32 & 35.01 & 40.89 & 35.03 & 35.47 \\
\midrule
\multirow{10}{*}{Multimodal} & BERT+ViT & 41.86 & 45.82 & 61.32 & 63.20 & 44.71 & 56.45 & 46.85 & 73.71 & 39.28 & 48.41 & 47.47 & 40.86 \\
& ViLT %\cite{40-kim2021vilt}
& 35.32 & 48.24 & 47.85 & 62.70 & 56.44 & 58.06 & 60.22 & 73.66 & 34.62 & 42.41 & 44.43 & 59.51 \\
& CLIP %\cite{41-radford2021learning}
& 53.22 & 65.83 & 63.65 & 70.93 & 67.17 & 67.43 & 70.86 & 79.06 & 44.99 & 59.86 & 55.29 & 40.98 \\
& TMPT %\cite{1-liang2024multi}
& 55.41 & 61.61 & 67.67 & \underline{76.60} & 63.19 & 67.25 & 62.92 & \underline{81.19} & 43.56 & 59.24 & 55.68 & 46.82 \\
& TMPT+CoT %\cite{1-liang2024multi}
& 66.61 & 68.75 & \underline{71.79} & {74.40} & \underline{69.96} & 68.43 & 63.00 & \textbf{82.71} & \underline{45.04} & \underline{60.52} & \underline{68.95} & \underline{59.87} \\
\cdashline{2-14}
& Qwen-VL %\cite{42-bai2023qwen}
& 43.31 & 45.13 & 50.51 & 43.06 & 45.49 & 49.79 & 46.04 & 27.73 & 36.50 & 40.78 & 42.14 & 39.34 \\
& MiMo  %\cite{44-xiaomi2025mimo} 
&66.21&65.89&61.54&45.32&42.19&50.69&42.36&40.26&39.30&58.95&60.07&52.39 \\
& LLaVa %\cite{45-li2024llava}
&49.48&45.80&60.30&20.48&28.48&33.83&38.94&31.57&36.53&47.47&50.45&47.07 \\
& Qwen2.5-VL %\cite{43-bai2025qwen2}
& 68.07 & 70.82 & 63.33 & 75.27 & 68.44& 67.15 &68.91 & 63.21 & 41.04 & 51.34 & 65.62 & 56.35 \\
& GPT4-Vision%\footnote{https://openai.com/research/gpt-4v-system-card}
& \underline{70.46} & \underline{72.82} & 61.63 & 44.59 & 47.07 & 57.47 & 57.90 & 37.61 & 44.83 & 56.40 & 66.72 & 56.90 \\

\midrule
\rowcolor{gray!20} 
& \textbf{MIND} & \textbf{70.87}$\uparrow$ & \textbf{73.31}$\uparrow$ & \textbf{72.90$^*$}$\uparrow$ & \textbf{78.49} $\uparrow$ & \textbf{71.85}$\uparrow$ & \textbf{71.89$^*$}$\uparrow$ & \underline{70.62}$\uparrow$ & 65.54$\uparrow$ & \textbf{50.81$^*$}$\uparrow$ & \textbf{62.42$^*$}$\uparrow$ & \textbf{69.38}$\uparrow$ & \textbf{61.13}$\uparrow$ \\
\multirow{3}{*}{Backbones}
&- w/ Qwen-VL&65.14$\uparrow$&63.84$\uparrow$&63.51$\uparrow$&55.76$\uparrow$&53.46$\uparrow$&50.67$\uparrow$&52.23$\uparrow$&50.06$\uparrow$&39.96$\uparrow$&49.30$\uparrow$&57.36$\uparrow$&52.75$\uparrow$\\
&- w/ MiMo&66.43$\uparrow$&69.91$\uparrow$&70.84$\uparrow$&59.25$\uparrow$&53.34$\uparrow$&61.26$\uparrow$&56.35$\uparrow$&58.18$\uparrow$&42.52$\uparrow$&60.22$\uparrow$&61.29$\uparrow$&54.82$\uparrow$\\
&- w/ LLaVa&58.26$\uparrow$&51.61$\uparrow$&62.38$\uparrow$&24.86$\uparrow$&28.88$\uparrow$&40.67$\uparrow$&43.14$\uparrow$&35.77$\uparrow$&36.79$\uparrow$&50.15$\uparrow$&51.81$\uparrow$&48.04$\uparrow$&\\

\bottomrule
\end{tabular}%
}
\caption{In-target results (\%). Best and second-best scores are in \textbf{bold} and \underline{underline}, respectively. $^*$ indicates our MIND model significantly outperforms baselines ($p < 0.05$). The dashed line separates fine-tuned from non-fine-tuned methods. w/ denotes the backbone, and $\uparrow$ indicates improvement over the original MLLM.}
\label{tab:in-target}
\end{table*}

\subsubsection{Stance Inference}
During the inference phase, MIND first leverages the SAM to extract visual information $\mathcal{V}$ from the input image, capturing fine-grained visual manifestations of semantic entities. This visual information is then paired with the text input $u$ to form a multimodal context.
It then employs its modality-synergistic retrieval mechanism to fetch the relevant prior experiences from both the modality and semantic experience pools. Finally, MIND integrates these experiences with multimodal context and the target to derive the final stance prediction, ensuring decisions are based on immediate data and guided by its refined long-term understanding.

\section{Experiments}
\subsection{Experiment Settings}
\textbf{Dataset and Metrics.}
To evaluate our MIND, we benchmark its performance on the MMSD dataset \cite{1-liang2024multi}, which comprises five distinct domains: Multi-modal COVID-CQ (MCCQ), Multi-modal Russo-Ukrainian Conflict (MRUC), Multi-modal Taiwan Question (MTWQ), Multi-modal Twitter Stance Election 2020 (MTSE), and Multi-modal Will-They-Won't-They (MWTWT). 
Following the MMSD dataset, we adopt the Macro-F1 score as the primary metric for evaluation.

\textbf{Implementation Details.}
Our MIND method is implemented upon the Qwen-2.5-VL-32B model as its backbone. For the experience pool retrieval, we set the coefficient $\alpha$ to 0.7, a relevance threshold $\tau$ to 0.7, and retrieve the top $k=3$ experiences to balance context with noise. 
Entity-related knowledge is sourced from Wikipedia using the Python Wikipedia library.

\subsection{Baseline Methods.}
We evaluate our method against three categories of baselines: 
\textbf{Text-only baselines:} 1) BERT \cite{33-devlin2019bert}, using uncased BERT-base model; 2) RoBERTa \cite{34-liu2019roberta}, using RoBERTa-base model; 3) KEBERT \cite{35-kawintiranon2022polibertweet};
4) LLaMA2 \cite{36-touvron2023llama}, using LLaMA2-70B-chat; 5) GPT4.
\textbf{Vision-only baselines:} 1) ResNet \cite{37-he2016deep}, using ResNet-50 v1.5; 2) ViT \cite{38-dosovitskiy2020image}, using ViT-base-patch16-224 model; 3) Swin Transformer (SwinT) \cite{39-liu2021swin}, using SwinV2-base-patch4-window12-192-22k model.
\textbf{Multimodal baselines:}  
1) ViLT \cite{40-kim2021vilt}, using vilt-b32-mlm model; 2) CLIP \cite{41-radford2021learning}, using CLIP-ViT-base-patch32 model; 3) BERT+ViT, a model that uses BERT and ViT as the text and image encoders;
4) Qwen-VL \cite{42-bai2023qwen}, using the Qwen-VL-Chat7B. 5) Qwen2.5-VL \cite{43-bai2025qwen2}, using the Qwen2.5-VL-32B-Instruct; 6) GPT4-Vision
; 7) MiMo \cite{44-xiaomi2025mimo}, using MiMo-7B-RL, 8) LLaVa \cite{45-li2024llava}, using LLaVa-v1.6-Mistral-7B; 9) TMPT and TMPT+CoT \cite{1-liang2024multi}.

\begin{table*}[t!]
\centering
\small
\setlength{\tabcolsep}{2.5pt}
% \resizebox{\textwidth}{!}{%
\begin{tabular}{ll ccccccccccc}
\toprule
\multirow{2.5}{*}{MODALITY} & \multirow{2.5}{*}{METHOD} & \multicolumn{2}{c}{MTSE}  & \multicolumn{4}{c}{MWTWT} & \multicolumn{2}{c}{MRUC} & \multicolumn{2}{c}{MTWQ} \\
\cmidrule(lr){3-4} \cmidrule(lr){5-8} \cmidrule(lr){9-10} \cmidrule(lr){11-12} 
& & DT & JB & CA & CE & AC & AH & RUS & UKR & MOC & TOC \\
\midrule
\multirow{5}{*}{Text-only} 
& BERT %\cite{33-devlin2019bert}
& 32.52 &29.97 &63.55 &61.30 &59.18 &52.89 &22.01& 15.45 &28.04 &9.57 \\
& RoBERTa %\cite{34-liu2019roberta}
&26.60& 32.21& 59.22& 59.22& 64.86& 57.46& 27.10& 19.98& 30.62& 15.84 \\
& KEBERT %\cite{35-kawintiranon2022polibertweet}
& 26.17& 31.81& 59.70& 62.56& 63.92& 55.53& 24.68& 28.18& 29.17& 19.80 \\
\cdashline{2-12}
& LLaMA2 %\cite{36-touvron2023llama}
& 53.57& 53.92 &32.47& 38.37& 48.08& 46.13& 31.86& 36.34& 51.46& 44.10\\
& GPT4%\footnote{https://openai.com/research/gpt-4} 
& 70.78& 68.83& 57.19& 60.56& 65.63& \textbf{69.01}& 40.22& 49.18& 62.10& 52.12 \\
\midrule
\multirow{3}{*}{Vision-only} & ResNet %\cite{37-he2016deep}
& 25.52& 29.70& 23.01& 24.11& 25.21& 25.27& 23.88& 25.57& 27.59& 24.88 \\
& ViT %\cite{38-dosovitskiy2020image}
& 28.63& 29.70& 24.59& 28.18& 34.06& 33.40& 27.26& 28.51& 29.37& 23.69 \\
& SwinT %\cite{39-liu2021swin}
& 28.54 &30.85& 28.53& 28.50& 35.87& 34.33& 25.44& 24.54& 27.90 &19.69 \\
\midrule
\multirow{10}{*}{Multimodal} & BERT+ViT & 26.70& 31.57& 59.21& 59.30& 65.04& 59.28& 23.33& 15.21& 24.76& 11.70 \\
& ViLT %\cite{40-kim2021vilt}
& 28.08 &29.74& 38.33& 46.00& 55.01& 48.55& 21.56& 23.96& 23.54& 19.18 \\
& CLIP %\cite{41-radford2021learning}
& 28.21 &28.99& 61.08& 55.67& 63.80& 60.06& 25.62& 27.40 &27.21& 15.69 \\
& TMPT %\cite{1-liang2024multi}
& 31.69 &32.65& 66.36& \underline{66.39}& \underline{66.32}& 61.56& 23.87& 24.71 &32.18 &26.48\\
& TMPT+CoT %\cite{1-liang2024multi}
& 54.30 &58.46& \underline{67.28}& 63.73 &64.87& 54.26& \underline{48.99}& \underline{51.75}& 45.32& 43.70\\
\cdashline{2-12}
& Qwen-VL %\cite{42-bai2023qwen}
& 47.62 &46.14& 38.57& 43.36& 47.82& 41.01& 36.95& 41.39& 44.32& 44.08 \\
& MiMo %\cite{44-xiaomi2025mimo}
& 66.58&66.01 &38.11& 40.10&44.20 &39.05 &44.85&50.24&64.65&46.25 \\
& LLaVa %\cite{45-li2024llava}
&50.26&46.74 &26.53 &30.77 & 34.96& 36.57&40.67&38.69&48.97&39.80 \\
& Qwen2.5-VL %\cite{43-bai2025qwen2}
& 68.11 &67.02& 62.59& 60.21& 62.52& 65.19& 48.22& 45.82& \underline{67.08}& 51.23 \\
& GPT4-Vision
 & \underline{72.68} &\underline{71.28}& 42.23& 45.92& 54.59& 53.19& 42.09 &47.00& 65.00& \underline{52.36} \\

\midrule
\rowcolor{gray!20}
& \textbf{MIND} & \textbf{73.78$^*$}$\uparrow$& \textbf{72.69$^*$}$\uparrow$& \textbf{70.83}$\uparrow$& \textbf{66.90}$\uparrow$ &\textbf{71.82$^*$}$\uparrow$& \underline{67.25}$\uparrow$& \textbf{50.07$^*$}$\uparrow$& \textbf{52.57$^*$}$\uparrow$& \textbf{69.77}$\uparrow$& \textbf{64.39$^*$}$\uparrow$ \\
\multirow{3}{*}{Backbones}
&- w/ QwenVL&59.48$\uparrow$&56.47$\uparrow$&59.71$\uparrow$&52.10$\uparrow$&58.09$\uparrow$&50.94$\uparrow$&44.12$\uparrow$&42.42$\uparrow$&60.16$\uparrow$&45.79$\uparrow$ \\
&- w/ MiMo&67.44$\uparrow$&66.09$\uparrow$&44.70$\uparrow$&40.12$\uparrow$&45.59$\uparrow$&41.27$\uparrow$&46.62$\uparrow$&52.23$\uparrow$&65.57$\uparrow$&46.47$\uparrow$\\
&- w/ LLaVa&53.10$\uparrow$&49.52$\uparrow$&27.21$\uparrow$&30.42&44.13$\uparrow$&38.25$\uparrow$&42.43$\uparrow$&40.64$\uparrow$&50.63$\uparrow$&45.76$\uparrow$
									
\\
\bottomrule
\end{tabular}%
% }
\caption{Zero-shot results (\%). Best and second-best scores are in \textbf{bold} and \underline{underline}, respectively. $^*$ indicates our MIND model significantly outperforms baselines ($p < 0.05$). The dashed line separates fine-tuned from non-fine-tuned methods. w/ denotes the backbone, and $\uparrow$ indicates improvement over the original MLLM.}
\label{tab:zero-shot}
\end{table*}

\subsection{Main Results}
\subsubsection{In-target Multimodal Stance Detection.}

As presented in Table \ref{tab:in-target}, our MIND achieves competitive performance across most baselines, even powerful LLMs such as GPT4 and GPT4-Vision in multiple scenarios. While TMPT+CoT shows stronger results on some targets, it relies on training with GPT4-generated CoT examples.
Instead, MIND highlights its superiority by leveraging accumulated experience for adaptive modality contribution rethinking.

A deeper analysis of baseline performance reveals the intricate challenges in multimodal fusion. The consistent outperformance of text-only models over vision-only counterparts confirms that text serves as the primary modality for stance determination. Yet, the superior performance of advanced multimodal models like GPT4-Vision compared to their text-only variants validates the supplementary value of visual cues, underscoring the necessity of cross-modality integration.
Notably, the inconsistent performance gains across fusion approaches, combined with the struggles of simpler frameworks like CLIP and TMPT, underscore a critical deficiency, as these models fail to handle the varying contribution of each modality on a per-sample basis, allowing noisy or irrelevant visual data to degrade overall performance. In contrast, MIND leverages its dual-reasoning paradigm to dynamically rethink modality contributions, thereby enhancing its final predictive accuracy.

\subsubsection{Zero-shot Multimodal Stance Detection.}
As shown in Table \ref{tab:zero-shot}, the results expose the inherent fragility of models reliant on in-domain training. While smaller architectures like BERT and RoBERTa suffer a precipitous performance collapse, even powerful LLMs such as GPT4 and GPT4-Vision demonstrate significant performance volatility when confronted with unseen targets. This underscores a fundamental limitation in current paradigms: true zero-shot generalization remains a formidable and largely unsolved challenge.
In this demanding scenario, MIND establishes a new standard for robustness and adaptability. It achieves state-of-the-art results on a majority of the evaluated datasets. This stability, particularly when contrasted with the erratic outputs of leading LLM-based baselines, provides compelling evidence of a superior generalization capability. 

\begin{table}[!t]
\centering
\small
\setlength{\tabcolsep}{5pt} 
 \resizebox{\linewidth}{!}{
\begin{tabular}{lccccc}
\toprule
METHOD & MTSE & MCCQ & MWTWT & MRUC & MTWQ \\
\midrule
\rowcolor{gray!20}
\multicolumn{6}{c}{\textit{\textbf{in-target setting}}} \\ %\cdashline{1-6}
MIND & \textbf{70.88} & \textbf{72.90} & \textbf{72.18} & \textbf{71.41} & \textbf{69.76} \\ \hdashline
-w/o MEP & 69.04 & 68.38 & 69.42 & 70.33 & 68.45 \\
-w/o SEP & 70.65 & 69.56 & 67.48 & 67.05 & 69.11 \\
-w/o CoT & 70.78 & 67.92 & 69.13 & 65.45 & 69.17 \\ \hdashline%\cdashline{1-7}
\rowcolor{gray!20}
\multicolumn{6}{c}{\textit{\textbf{zero-shot setting}}} \\ %\cdashline{1-7}
MIND & \textbf{71.19}&- & \textbf{70.59} & \textbf{70.74} & \textbf{71.39} \\  \hdashline
-w/o MEP & 66.95& -& 63.26 & 65.71 & 68.55 \\
-w/o SEP & 64.47&- & 64.35 & 68.74 & 70.41 \\
-w/o CoT & 70.85& -& 64.77 & 69.11 & 71.22 \\

\bottomrule
\end{tabular}
}
\caption{Experimental results (\%) of the ablation study
 on in-target and zero-shot multimodal stance detection.}
\label{tab:ablation}
\end{table}

\subsection{Ablation Study}
To evaluate the impact of each component within our MIND, we conduct an ablation study, with results reported in Table \ref{tab:ablation}. The findings unequivocally demonstrate the synergistic importance of all modules, as the full MIND model consistently outperforms any ablated version across both in-target and zero-shot settings. The most critical components are the experience pools. Without using modality experience (w/o MEP) results in the most significant performance degradation, particularly in the zero-shot scenario, highlighting the indispensable role of assessing cross-sample modality contribution for robust generalization. Similarly, the exclusion of using semantic experience (w/o SEP) causes a severe decline, confirming the necessity of leveraging semantic context for accurate stance inference. Finally, while removing the CoT (w/o CoT) module also impairs performance, the impact is less pronounced, suggesting that CoT functions as a higher-level reasoning synthesizer, which organizes and refines the foundational evidence provided by the MEP and SEP to formulate a coherent final stance.

\subsection{Effect of Different Backbones}
We assess the performance of MIND with different MLLM backbones in both in-target and zero-shot settings, as detailed in Table \ref{tab:in-target} and Table \ref{tab:zero-shot}. 
Smaller backbones generally yield weaker performance, likely due to their limited visual perception capabilities. 
However,  integrating any MLLM as MIND's backbone still brings significant improvements marked by $\uparrow$ compared to standalone operation, with Qwen-VL showing particularly notable gains.
This consistent enhancement across diverse architectures highlights that our framework effectively handles varying modality contributions by dynamically weighting them based on their actual expressive power for target stances. It thus compensates for limitations in smaller backbones, ensuring robust performance regardless of the backbone's inherent biases toward text or vision.

\begin{figure}[!t]
  \centering
  \includegraphics[width=0.95\linewidth]{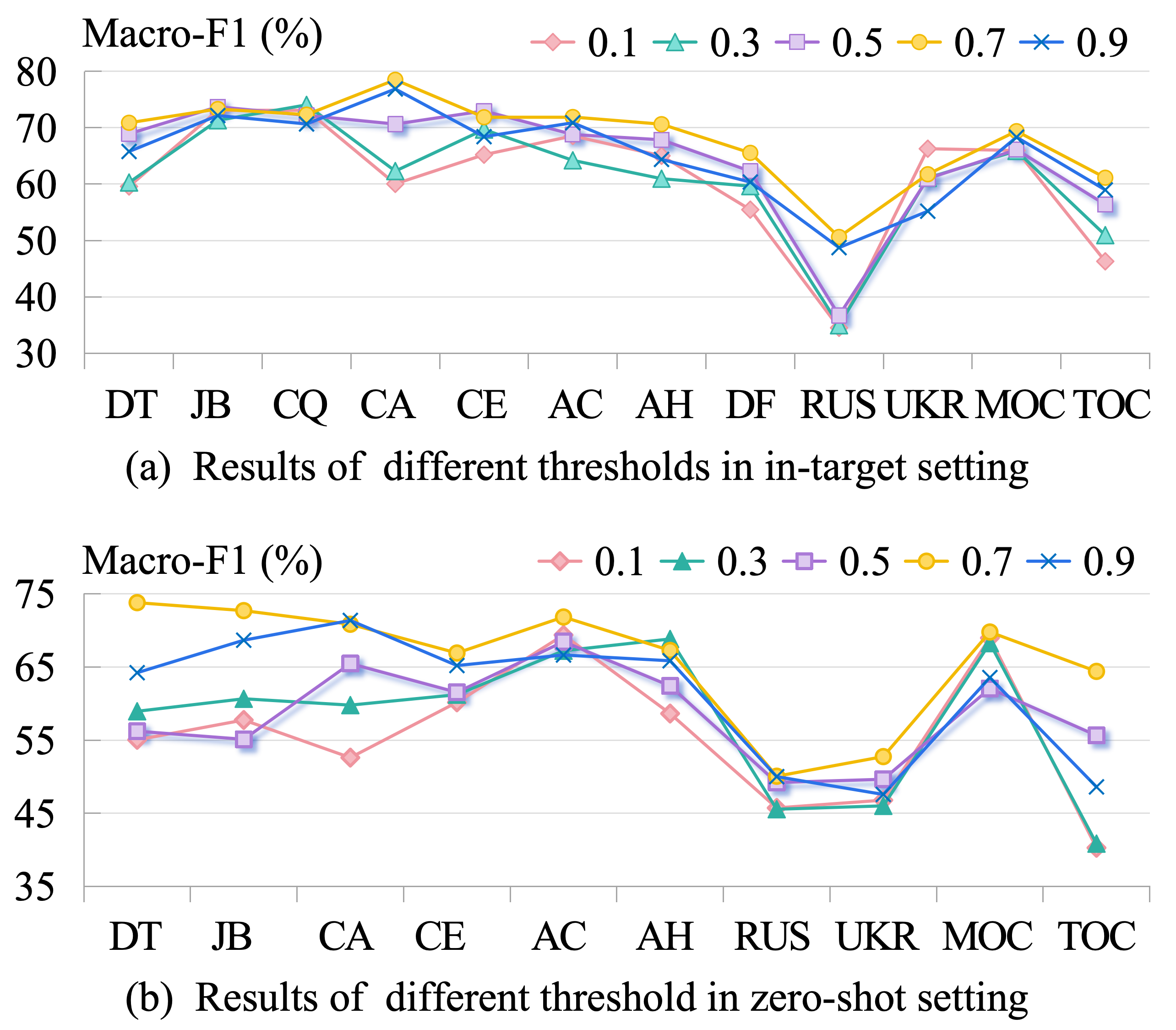}
  \caption{Results on different relevance thresholds.}
  \label{threshold}
\end{figure}

\subsection{Different Relevance Thresholds}
We analyzed the effect of varying relevance thresholds (0.1, 0.3, 0.5, 0,7, and 0.9) on retrieval performance for both in-target and zero-shot settings, as shown in Figure \ref{threshold}.
The results indicate that in both settings, performance tends to be weaker at lower thresholds and gradually degrades once the threshold exceeds 0.7.
This trend highlights a critical trade-off between the quality and quantity of retrieved experiences. 
A low threshold risks retrieving irrelevant or poor-quality experiences, which can introduce noise and impair the reasoning process. Conversely, an excessively high threshold becomes too restrictive, retrieving an insufficient number of experiences and thus limiting the model's ability to leverage its accumulated knowledge for robust inference. 

\begin{figure}[!t]
  \centering
  \includegraphics[width=\linewidth]{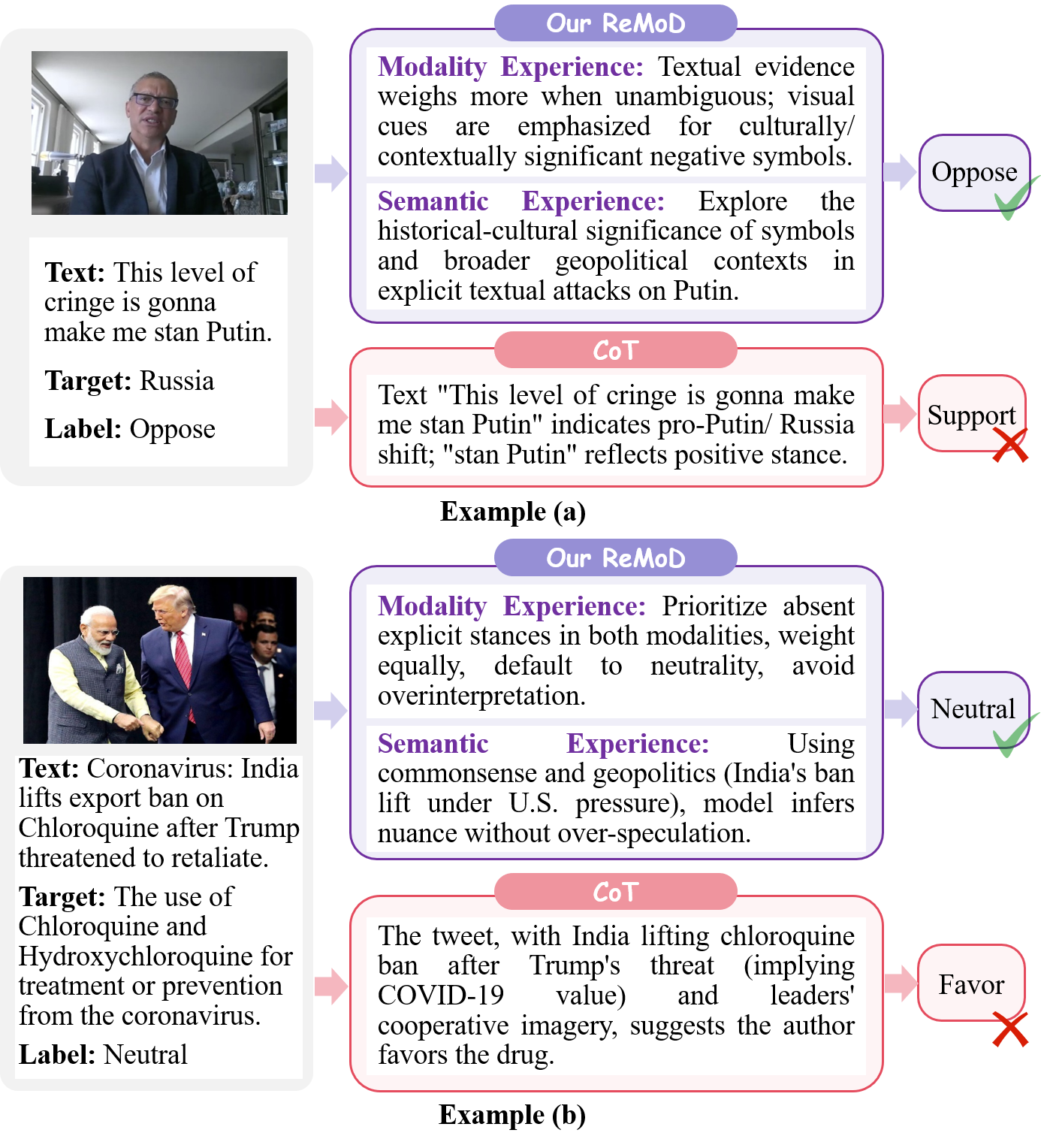}
  \caption{Case study between our MIND and directly CoT method based on the same LLM backbone.}
  \label{case_study}
\end{figure}

\subsection{Case Study}
To further underscore the strengths of MIND’s dual reasoning, we analyze two examples in Figure \ref{case_study} by comparing the results of direct CoT with MIND.
In Example (a), direct CoT literally interprets `stan Putin' in the text as support but ignores the negative context of `cringe', which renders `stan' sarcastic. MIND’s modality experience emphasizes weighting textual evidence via contextual judgment, enabling recognition of implicit sarcasm through the contradiction between \textit{cringe} and \textit{stan} to avoid being misled by isolated explicit terms. 
Its semantic experience clarifies that such sarcasm criticizes Russia-related phenomena, thus linking the text to the target stance of opposing Russia. 

For Example (b), direct CoT infers support for the drug from surface information like India lifting the export ban in text and leaders’ collaboration in images. In contrast, MIND’s modality experience stresses prioritizing explicit stance information across modalities, equal weighting, and default neutrality, while it deepens understanding via common sense and geopolitical context, revealing that the text describes objective policy adjustments amid international games rather than the author’s subjective endorsement of the drug’s efficacy. By curbing surface information misinterpretation and calibrating reasoning with background knowledge, MIND accurately captures the neutral stance.

\section{Conclusion}
This paper proposes MIND, a framework for multimodal stance detection that employs a dual-reasoning paradigm inspired by human cognition. By integrating experience-driven intuitive reasoning with deliberate reflective reasoning, it dynamically captures the variable contributions of different modalities, overcoming the limitations of fixed fusion strategies. The dual-pool mechanism enables the accumulation and refinement of modality and semantic insights, enhancing robustness and context awareness. Experiments on MMSD show MIND outperforms most baselines in both in-target and zero-shot settings, with strong generalization.

\section*{Limitations}
A primary limitation of this work is its computational cost and inference latency. To emulate the complex dual-reasoning process of human cognition, our MIND framework involves multiple calls to the large multimodal model at various stages, including knowledge perception, intuitive reasoning, and reflective reasoning. While this deliberate design is crucial for achieving dynamic and adaptive reasoning, it inevitably leads to significant computational overhead, making it less suitable for real-time applications. Future work could explore techniques such as knowledge distillation or model compression to create a more lightweight version of the model while preserving its core reasoning capabilities.

\bibliography{custom}

% \appendix

\end{document}